\documentclass[letterpaper, 10 pt, conference]{ieeeconf}  %

\IEEEoverridecommandlockouts                              %

\usepackage{amsmath}   %
\usepackage{amssymb}   %
\usepackage{amsfonts}
\usepackage{bm}
\usepackage{psfrag}
\usepackage{color}
\usepackage{epic,bez123}
\usepackage{floatflt}
\usepackage{multirow}
\usepackage{wrapfig}
\usepackage{graphicx}
 \usepackage{epstopdf}
\usepackage{algorithmicx}
\usepackage[ruled]{algorithm}
\usepackage{listings}
\usepackage{subfig}

\usepackage{algpseudocode}
\usepackage{tikz}

\title{\LARGE \bf
	Towards Long-endurance Flight: Design and Implementation of a Variable-pitch Gasoline-engine Quadrotor
}

\author{Tao Pang$^{1}$, Kemao Peng$^1$, Feng Lin$^1$ and Ben M. Chen$^{2}$%
	\thanks{$^{1}$Tao Pang, Kemao Peng and Feng Lin are with the Temasek Laboratories, NUS, Singapore.
		{\tt\small tslpt@nus.edu.sg, linfeng@nus.edu.sg, kmpeng@nus.edu.sg}}%
	\thanks{$^{2}$Ben M. Chen is with the Department of Electrical \& Computer Engineering, NUS, Singapore.
		{\tt\small bmchen@nus.edu.sg}}%
}

\begin{document}
	
	\maketitle
	\thispagestyle{empty}
	\pagestyle{empty}

	\begin{abstract}
		
	Majority of today's fixed-pitch, electric-power quadrotors have short flight endurance ($<$ 1 hour) which greatly limits their applications. This paper presents a design methodology for the construction of a long-endurance quadrotor using variable-pitch rotors and a gasoline-engine. The methodology consists of three aspects. Firstly, the rotor blades and gasoline engine are selected as a pair, so that sufficient lift can be comfortably provided by the engine. Secondly, drivetrain and airframe are designed. Major challenges include airframe vibration minimization and power transmission from one engine to four rotors while keeping alternate rotors contra-rotating. Lastly, a PD controller is tuned to facilitate preliminary flight tests. The methodology has been verified by the construction and successful flight of our gasoline quadrotor prototype, which is designed to have a flight time of 2 to 3 hours and a maximum take-off weight of 10 kg.
	
	\end{abstract}

	\section{Introduction \label{sec.Intro}}
	Primarily due to their mechanical simplicity, quadrotor unmanned aerial vehicles (UAVs) have become a popular platform for research and civilian activities in recent years. Various research groups have used quadrotor UAVs to implement algorithms in many research areas, such as vision/laser based navigation and simultaneous localization and mapping (SLAM). On the other hand, quadrotors carrying digital cameras are also widely used in aerial photography, power line inspection and many other situations. 
	
	Quadrotors today are almost exclusively battery-powered, and consist of four electric motors, each of which directly drives a fixed-pitch propeller. Accordingly, the motion of the quadrotor is controlled by differential motor speed. While this simple configuration is the reason behind quadrotors’ mechanical simplicity, it is also the reason for one of quadrotors’ major drawbacks: limited flight endurance. According to technical specifications provided by major quadrotor manufacturers, flight endurance of most quadrotors today ranges from 20 minutes to an hour, which greatly limits their practical applications. 
	
	It is possible, however, to increase the flight endurance of quadrotors to a few hours by using a gasoline engine, and employing an appropriate drivetrain to transmit power from the engine to the rotors. However, control actions can no longer be generated by differential rotor speed because all rotors, which are mechanically linked by the drivetrain, rotate at the same speed. In order to generate the necessary control actions, variable-pitch propellers have to be used, which provide varying torque and lift by changing the pitch angle of rotor blades. In addition, variable-pitch propellers also have the advantage of better agility and higher control bandwidth \cite{cutler2011comparison}. Although mechanical complexity is ramped up by the drivetrain and variable-pitch mechanism, gas-engine-powered quadrotors are an interesting and attractive alternative to battery-powered quadrotors, due to the potential of increased flight endurance and enhanced agility.

	\begin{figure}[tp]
		\centering
		\includegraphics[width=84mm]{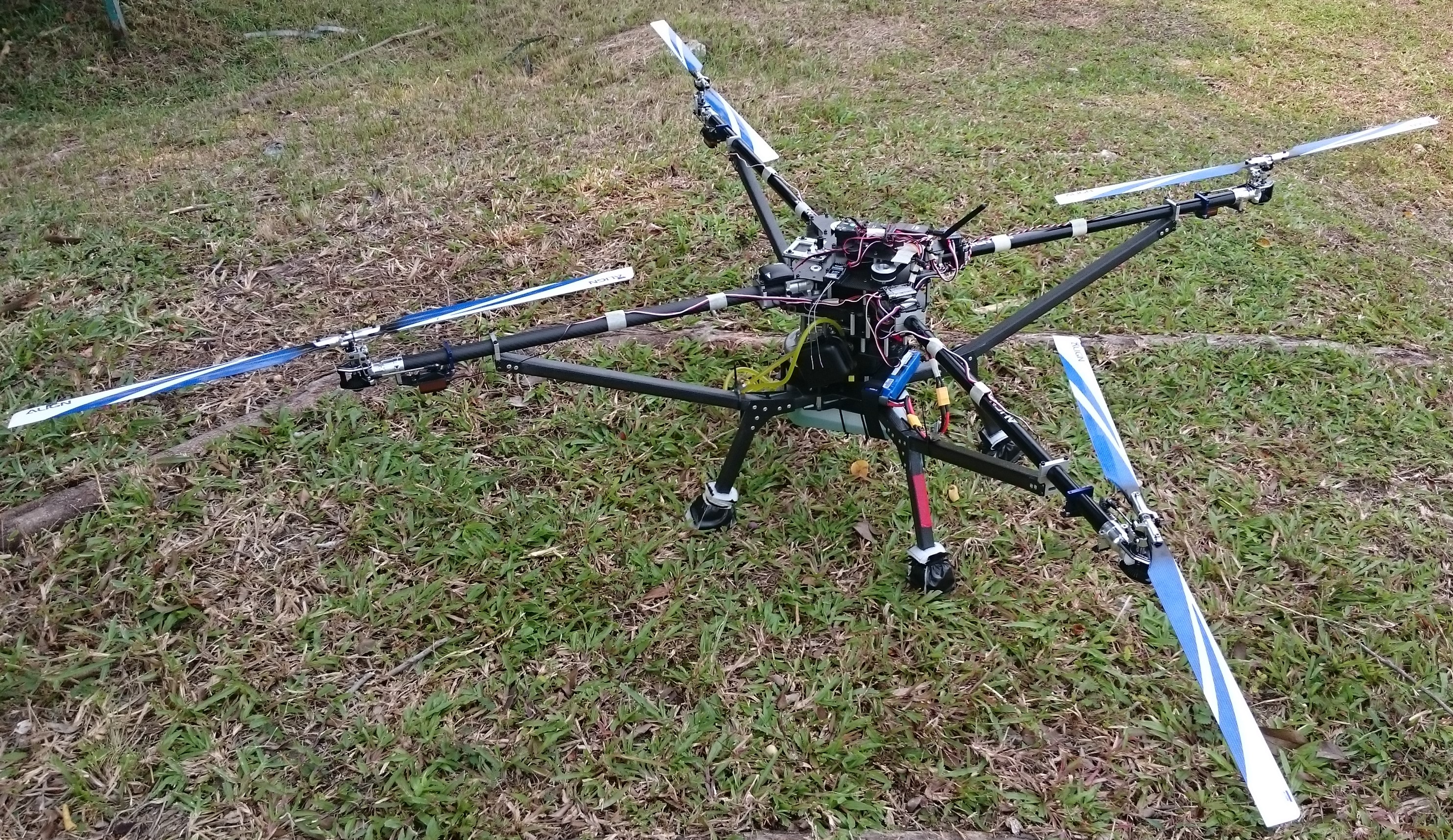}
		\caption{Completed gasoline quadrotor prototype}
		\label{fig:full_prototype}
	\end{figure}
	
	Variable-pitch, gasoline-engine quadrotor prototypes have been built by a few groups of hobbyists. Although some of them are known to be able to fly, none has demonstrated the reliability necessary for long-endurance flight \cite{GasQuad_DIYDrones}\cite{HG3}\cite{Stingray500}. On the other hand, the idea of variable-pitch quadrotors has been explored by a handful of researchers to improve quadrotors' agility, but their prototypes are too small to carry a gasoline engine \cite{kawasaki2013muwa}\cite{michini2011design}. Moreover, issues vital to gasoline engine quadcopters, such as drivetrain and airframe design, have not been addressed in design methodologies of their smaller electric counterpart, whose main focuses are usually aerodynamics, dynamic modeling and control strategy \cite{hoffmann2007quadrotor}\cite{pounds2009design}.
	
	In this paper, we propose a methodology to design and build variable-pitch gasoline-engine quadrotors, with the intention to increase quadrotors' flight endurance. Using our quadrotor prototypes as examples, the methodology aims to highlight essential design considerations as well as address issues overlooked in conventional quadrotor design. Section \ref{sec.overview} gives an overview of the methodology and our prototypes. Section \ref{sec.porp_engine} explains how requirements on lift, power and flight time can be achieved by the selection of rotor blades and engine. Section \ref{sec.Drivetrain_Airframe} introduces how to construct the drivetrain by creatively using mechanical components, and how to minimize airframe vibration by analysis and experiments. Section \ref{sec.flight control} describes flight control of our gasoline quadrotor prototypes. Section \ref{sec.flight tests} shows flight test results and Section \ref{sec.concl} discusses conclusions and future works.

	\section{Design Overview \label{sec.overview}}
	Our gasoline quadrotor (Figure \ref{fig:full_prototype}) resembles conventional X-shape quadrotors. The rotors are located at the tips of the ``X" with the engine, avionics, batteries, fuel tank and additional sensors stacked at the center.  
	\begin{figure}[htp]
		\centering
		\includegraphics[width=85mm]{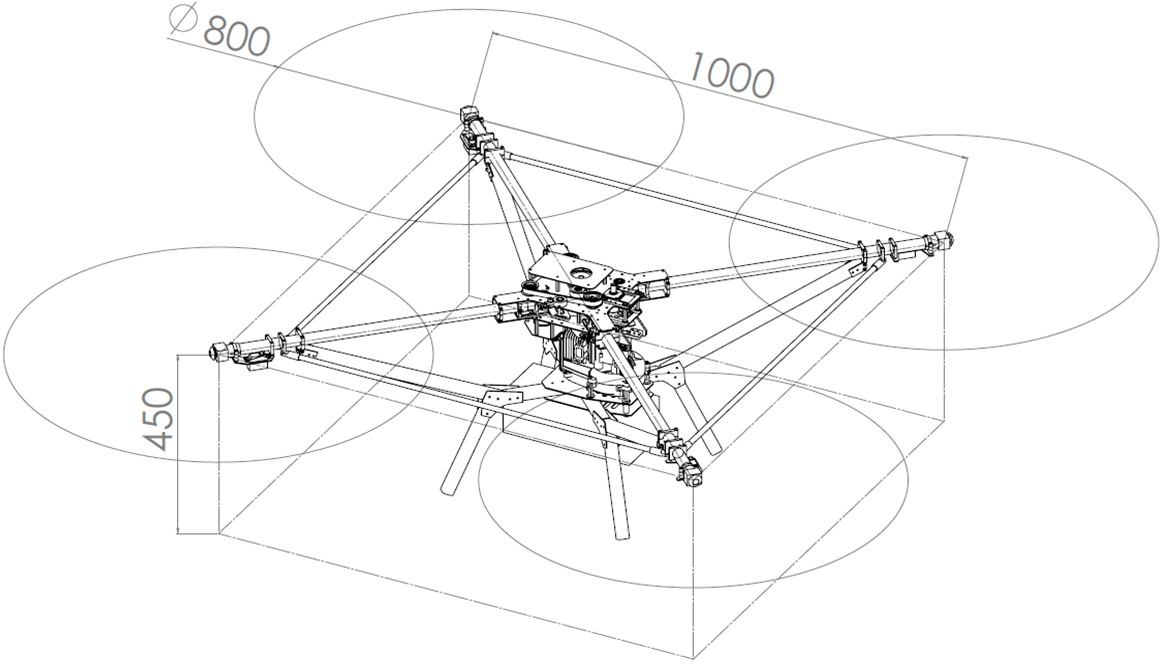}
		\caption{Gasoline quadrotor dimensions (mm)}
		\label{fig:gasoline_overview}
	\end{figure} 
	
	The first step in the design of the gas-engine quadrotor was to determine the size of the rotors and the engine, which determined its size and weight. The drivetrain and airframe were then designed accordingly. Finally, the quadrotor was wired up to a commercial autopilot named Pixhawk. A PD controller was used for attitude control, whose gains can be tuned during flight tests.
	
\begin{figure}[htp]
		\centering
		\includegraphics[width=80mm]{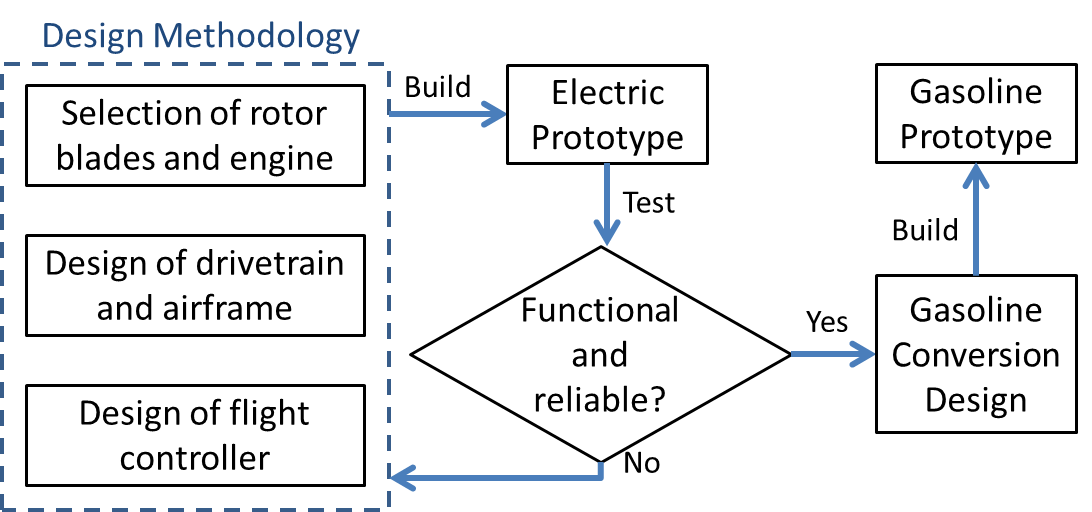}
		\caption{Design flowchart}
		\label{fig:methodology_flowchart}
	\end{figure}

	It is well known that compared with electric motors, gasoline engines require a more complicated drivetrain and induce more vibration to the airframe. Therefore, a proof-of-concept prototype powered by an electric motor was first built to verify key aspects of the design methodology before tackling the additional complication of gasoline engines. After the electric prototype proved to be successful, we proceeded with the conversion from electricity to gasoline, which mainly involved the addition of a clutch mechanism and reinforcement of the airframe. A schematic diagram of this design and testing process is shown in Figure \ref{fig:methodology_flowchart}.
	
	Powered by a 1.8 kW engine, the gasoline quadrotor has a maximum thrust of 16 kg, a maximum take-off weight of 10kg and an empty weight of 6.5 kg. Dimensions of the gasoline prototype is shown in Figure \ref{fig:gasoline_overview}. Having the same propulsion capacity and dimensions as the gasoline quadrotor, the electric prototype uses a motor with a maximum output of 2 kW powered by 22.2 V 6s Li-po batteries.

	\section{Selection of Rotor and Engine \label{sec.porp_engine}}
	As customization of rotor blade and engine can be complicated and time-consuming \cite{pounds2009design}\cite{pounds2004towards}, off-the-shelf hobby helicopter components are used for the implementation of our prototypes. A survey of hobby gas engines and rotor blades reveals that the Zenoah gas engine series with an output from 1.5 to 2.5 kW is mature, reliable and widely used for small-scale gasoline helicopters. On the other hand, hobby helicopter main rotor blades whose diameters range from 0.3 to 1.8 m can serve as the blades for our gasoline quadrotor.
	
	The final rotor and engine combination depends on how much power is consumed by a rotor of a certian size, which can be estimated using momentum theory and blade element theory. For a single rotor, momentum theory estimates the minimum power required to generate a certain lift $N$, which is given by 
	
	\begin{equation}
	P=\frac{N^{\frac{3}{2}}}{\sqrt{2{\rho}A}}
	\end{equation}
	where $P$ is the total power, $N$ is the total lift, $\rho$ is the density of air and A is the rotor area \cite{johnson2012helicopter}. If we consider a multirotor generating a total lift $N$ with $n$ rotors, each having a disk area $A_m$, the total power consumed by the multirotor is given by 
	
	\begin{equation}
	P_m=n\frac{{\left(\frac{N}{n}\right)}^\frac{3}{2}}{\sqrt{2{\rho}A_m}}=\frac{N^\frac{3}{2}}{\sqrt{2{\rho}(nA_m)}}
	\end{equation}
	
	It can be seen that for a given lift, the total power consumed by a multirotor is inversely proportional to the square root of the total rotor area, $nA_m$. The conclusion given by momentum theory, therefore, is that power per unit lift can be minimized by maximizing total rotor area.
	
	The relationship between lift, power and the pitch angle of the rotor can be further deduced from blade element theory. The lift and power coefficients are defined respectively as 
	\begin{equation}
	\label{eqn:definition of lift coefficient}
	C_T=T/{\rho}A{({\Omega}R)}^2
	\end{equation}
	
	\begin{equation}
	\label{eqn:definition of torque coefficient}
	C_P=P/{\rho}A{({\Omega}R)}^3
	\end{equation}	
	where $\Omega$  and $R$ are the angular velocity and radius of the rotor respectively, and $A$ is the rotor disk area \cite{johnson2012helicopter}.
	
	It can be shown that 
	\begin{equation}
	\label{eqn:lift_coefficient}
	C_T=\frac{{\sigma}a}{2}\left(\frac{\theta}{3}-\frac{\lambda}{2}\right)
	\end{equation}
	
	\begin{equation}
	\lambda=\frac{{\sigma}a}{16}\left[\sqrt{1+\frac{64}{{\sigma}a}\theta}-1\right]
	\end{equation}
	where $\lambda$ is called the inflow ratio, $\sigma$ is the rotor solidity ratio defined as the ratio of the rotor surface area over the rotor disk area, $C_l=a\alpha$ where $C_l$ is the sectional lift coefficient and $\alpha$ is the angle of attack \cite{johnson2012helicopter}. 
	
	The power coefficient $C_P$ is given by
	
	\begin{equation}
	C_P= C_{P_i} + C_{P_o}
	\end{equation}
	
	\begin{equation}
	C_{P_i}=\kappa\frac{{C_T}^{\frac{3}{2}}}{\sqrt{2}}
	\end{equation}

	\begin{equation}
	\label{eqn:power_coefficient}
	C_{P_o}=\frac{\sigma}{8}\left({\beta}_0+{\beta}_1\alpha+{\beta}_2{\alpha}^2\right)
	\end{equation}
	where $C_{P_i}$ stands for induced power (to generate lift) with $\kappa$ being the empirical induced power correction factor, and $C_{P_o}$ for profile power (to counter drag) with ${\beta}_1$, ${\beta}_2$ and ${\beta}_3$ being empirical coefficients \cite{johnson2012helicopter}. Values of the aerodynamic constants are shown in Table \ref{tab:aero_constants}.
	\begin{table}[h]
	\caption{Aerodynamic constants}
	\label{tab:aero_constants}
	\begin{center}
		\begin{tabular}{cccccc}
			\hline
			$a$ & $\kappa$ & ${\beta}_0$ & ${\beta}_1$ & ${\beta}_2$ & $\rho$ \\
			\hline
			5.7 & 1.60 & 0.0130 & -0.0216 & 0.400& 1.18$kg/m^3$ \\
			\hline
		\end{tabular}
	\end{center}
	\end{table}

\begin{figure}[h!]
		\centering
		\subfloat[Lift vs. pitch angle]{
			\includegraphics[width=88mm]{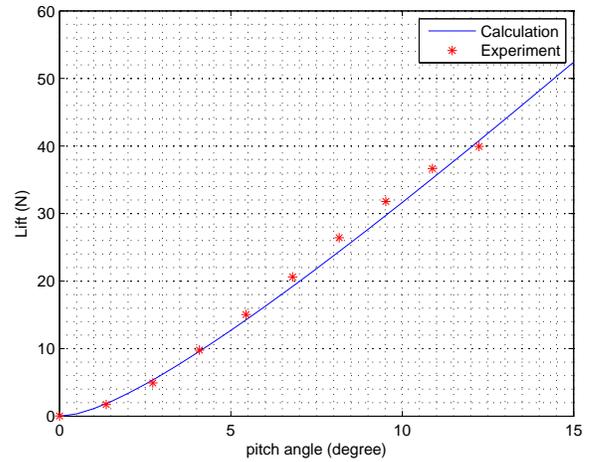}
			\label{fig:lift}
		}
		\\
		\subfloat[Power and torque vs. pitch angle]{
			\includegraphics[width=88mm]{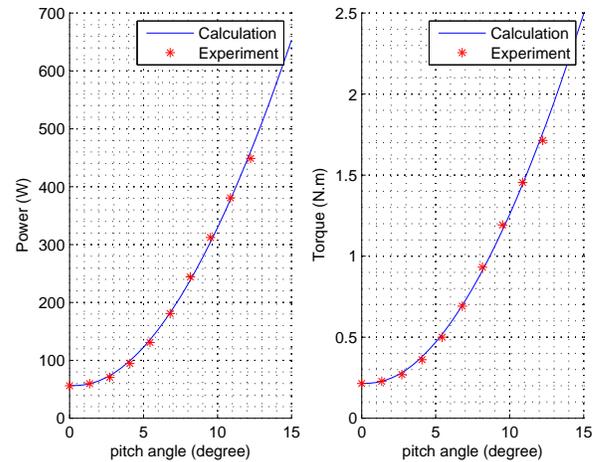}
			\label{fig:power_torque}
		}
		\caption{Lift, Power and torque of 800 mm rotor at 2500 RPM}%
		\label{fig:lift_power_torque}
	\end{figure}

	Lift, drag and power consumption for rotor blades of different sizes can thus be calculated using equation \ref{eqn:definition of lift coefficient} to \ref{eqn:power_coefficient}. After examining several hobby rotor blades, it has been found that the 800 mm diameter rotor blades used for 450 class hobby helicopters generate sufficient lift while consuming a reasonable amount of power. 800 mm is also the largest rotor diameter the airframe can accommodate given other design constraints. Using the aforementioned equations, the relationship between pitch angle and lift, power and torque at 2500 revolution per minute (RPM), the operating RPM recommended by blade manufacturer, is calculated and shown in Figure \ref{fig:lift} and \ref{fig:power_torque}.

	\begin{figure}[h!]
		\centering
		\includegraphics[width=80mm]{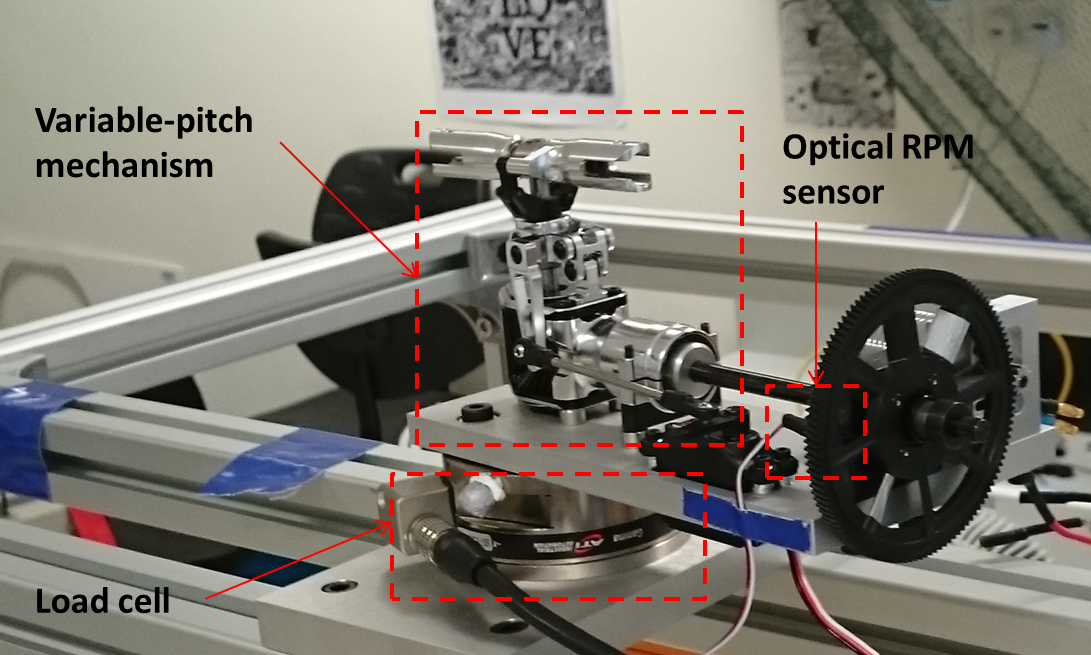}
		\caption{Single rotor lift and torque test stand}
		\label{fig:single_rotor_test_stand}
	\end{figure}
	
	The results from theoretical calculations were further corroborated by experiments. A test stand consisting of a load cell, an optical RPM sensor and the same variable-pitch mechanism used on the gasoline quadrotor was built to measure the thrust, torque and rotational speed of the  rotor at different pitch angles (Figure \ref{fig:single_rotor_test_stand}). As shown in Figure \ref{fig:lift} and \ref{fig:power_torque}, the measured torque and lift (shown as asterisks) agree well with the theoretical results. 

	The maximum lift, maximum take-off weight and engine horsepower can thus be determined using the results above. The operating range of pitch angle for the 800mm rotor is between $\pm14^{\circ}$. As seen in Figure \ref{fig:lift}, the lift generated by one rotor is 39 N, or 4.0 kg when pitch angle is equal to $14^{\circ}$, giving a total maximum lift of 16 kg. Using a 50\% control margin, the maximum take-off weight of the quadrotor can be as much as 10 kg. It is also observed in Figure \ref{fig:power_torque} that the total power and torque at maximum pitch are 1.3 kW and 5.2 Nm respectively. Based on the power and torque requirements, the Zenoah 270RC single cylinder gasoline engine with a displacement of 25.4 cc is chosen. Detailed torque and power curves of the engine can be found in \cite{Zenoah270RC}. With proper gearing, the engine is able to provide sufficient power and torque for the gasoline power quadrotor. 
	
	Flight endurance can also be estimated using Figure \ref{fig:lift_power_torque}. It is known that the empty weight of the gasoline quadrotor is 6.5 kg (Section \ref{sec.overview}), and that fuel consumption for Zenoah RC series engine is 554 g/kWh \cite{Zenoah270RC}. At maximum take-off weight (10 kg), the pitch angle required to take off is  $8^{\circ}$ with the quadrotor consuming 1.0 kW in total. Assuming that 2.5 L of gasoline is carried on-board and a transmission efficiency of 80\%, the flight time can be roughly estimated as

	\begin{equation}
	\frac{2.5L \times 770g/L}{554g/kWh {\times }(1kW / 0.8)} = 2.8\:\textnormal{hours} \\
	\end{equation}

	\section{Drivetrain and Airframe Design \label{sec.Drivetrain_Airframe}}	
	\subsection{Drivetrain \label{subsec.Drivetrain}}
\begin{figure}[h!]
		\centering
		\includegraphics[width=86mm]{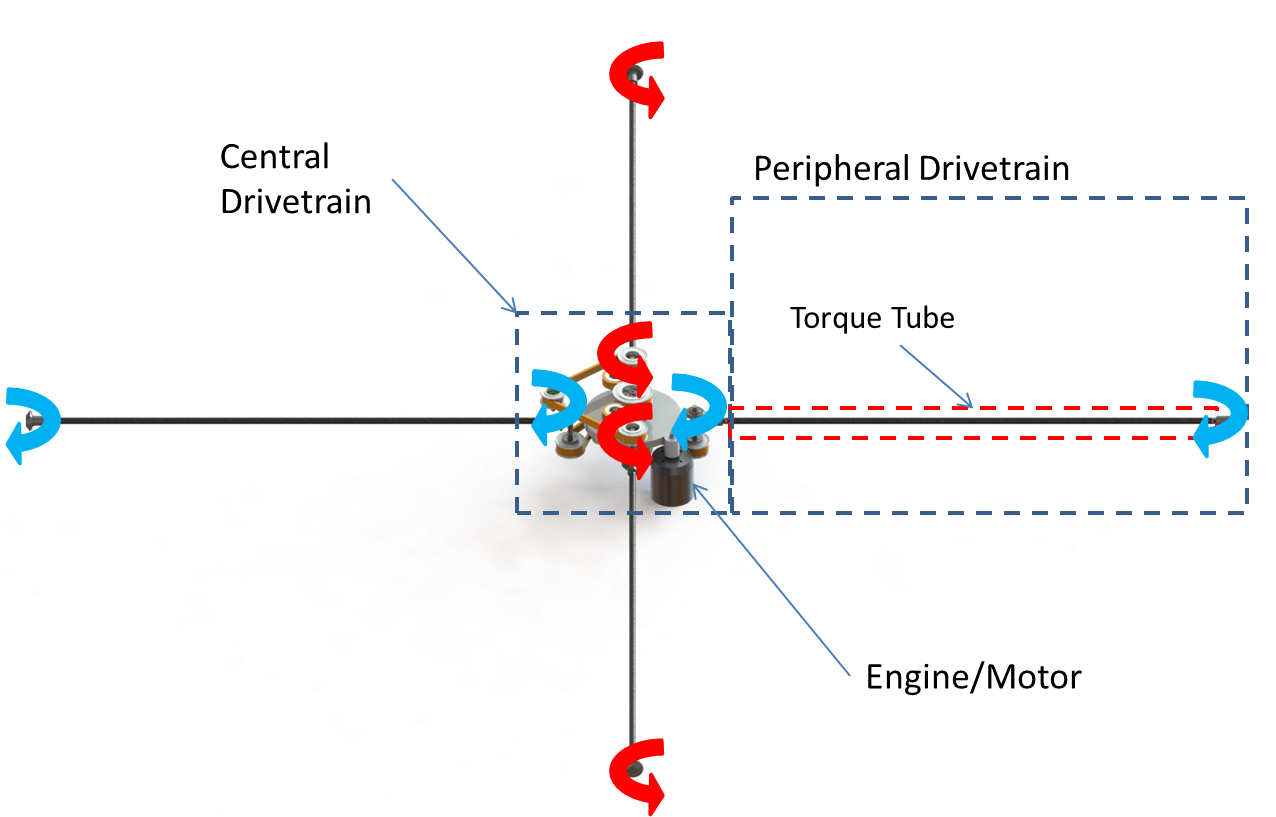}
		\caption{The entire drivetrain, red arrow indicates counter-clockwise rotation and blue clockwise rotation}
		\label{fig:drivetrain}
	\end{figure}

	The drivetrain is responsible for transmitting power from the engine to the rotors. In our design, the drivetrain can be divided into two subsystems. One is the central drivetrain, a combination of gears and belts that converts the rotation of the engine to four rotational outputs with alternate outputs rotating in opposite directions (arrows in Figure \ref{fig:drivetrain}). The other, referred to as the peripheral drivetrain, consists of a shaft (known as torque tubes in hobby terminology) and bevel gears which transmits power from the central drivetrain to the rotors. The peripheral drivetrain utilizes the same transmission mechanism as hobby helicopter tail rotors. The drivetrain design had gone through iterations of modification and tests before the finalized version was adopted for our prototypes.
	
\begin{figure}[h!]
		\centering
		\includegraphics[width=86mm]{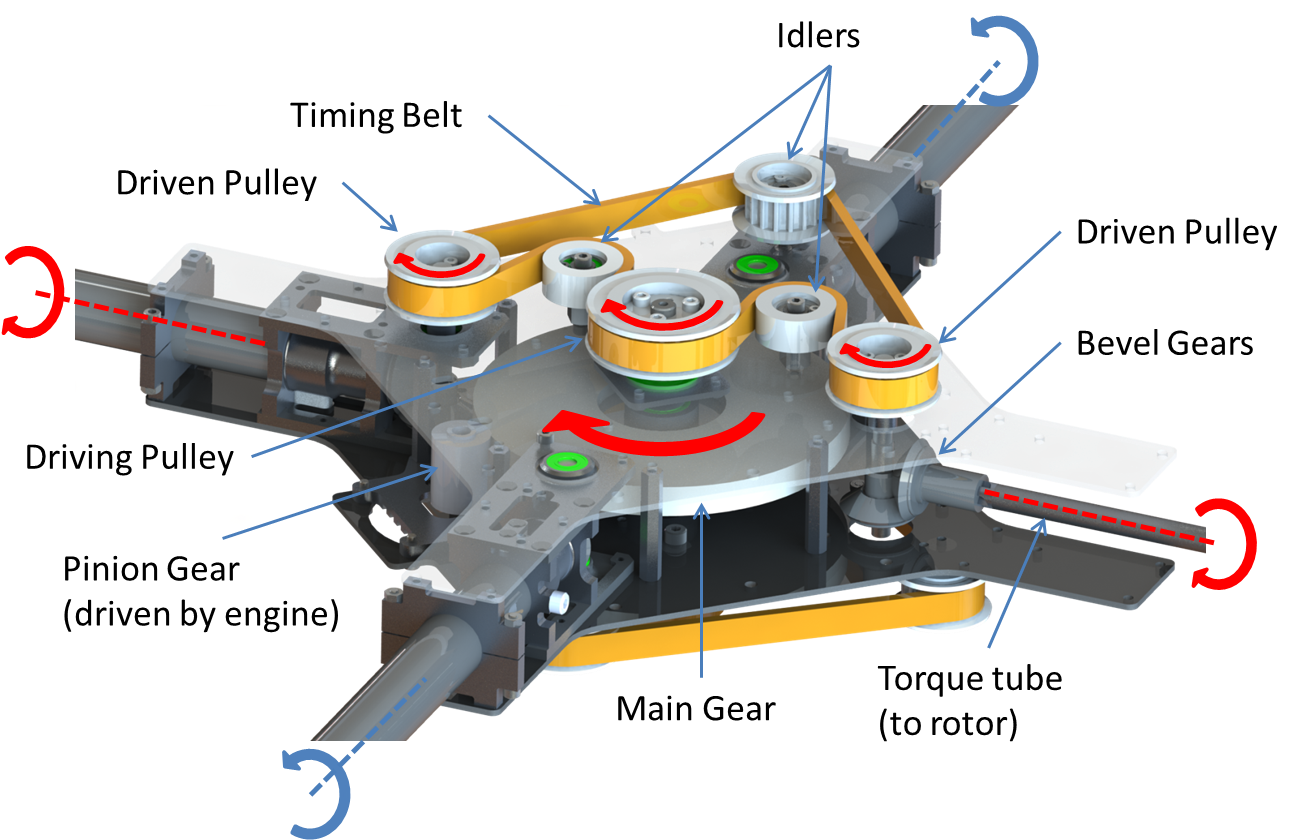}
		\caption{The central drivetrain}
		\label{fig:drivetrain_central}
	\end{figure}

	A detailed diagram of the central drivetrain is shown in Figure \ref{fig:drivetrain_central}. To make alternate rotors contra-rotating, the central drivetrain employs two decks of identical belt transmission systems with the bottom deck driving rotors in the clockwise direction, and the top deck the counter-clockwise direction. The top deck's driving pulley drives two driven pulleys, which in turn drives the torque tubes via a pair of bevel gears.  The torque tubes driven by the upper deck ultimately spin the counter-clockwise rotors. On the other hand, the occluded bottom deck is identical to the top deck but turned upside down to reverse its output direction. Both decks are driven by a shaft connected to the main gear, which is driven by the pinion gear connected to the engine. 

	It should be noted that other drivetrain designs are also possible, some of which have already been implemented by hobbyists \cite{GasQuad_DIYDrones}\cite{HG3}. Our drivetrain design may not be as good as existing ones in some aspects. For instance, our drivetrain is heavier and more complex than the bevel-gear-based drivetrain used in \cite{GasQuad_DIYDrones}. Nevertheless, our design featuring a combination of belts and gears has several unique advantages. First, belts allow greater freedom to place components, are less noisy and do not require lubrication. In addition, the belt-based central drivetrain can be conveniently scaled-up to hex or even octrotor by rerouting the belts, which is not as easy for other known drivetrains.

	\subsection{Airframe \label{subsec.Airframe}}
	Vibration of airframe becomes an prominent issue for quadrotors of our size (Figure \ref{fig:gasoline_overview}). A simple way to reduce vibration would be to strengthen the conventional X-shape quadrotor frame with stronger structural members. However, due to the availability of and compatibility between components, our prototype's airframe has to be made of slender and long structural members. Moreover, imprudent use of heavy structural members imposes a penalty on the quadrotor's payload capacity. As a result, the airframe has to be carefully designed and analyzed to minimize vibration and weight.
	
	It is known that the greatest contributor to helicopter vibration is the n-per-revolution vibration of the main rotor, where n is the number of blades of a single rotor. In our case, this vibration frequency can be calculated as
	\begin{equation}
	2500\:\textnormal{RPM} / 60 \times 2 = 83.3\:\textnormal{Hz}
	\end{equation}	
	
\begin{figure}[htp]
		\centering
		\includegraphics[width=88mm]{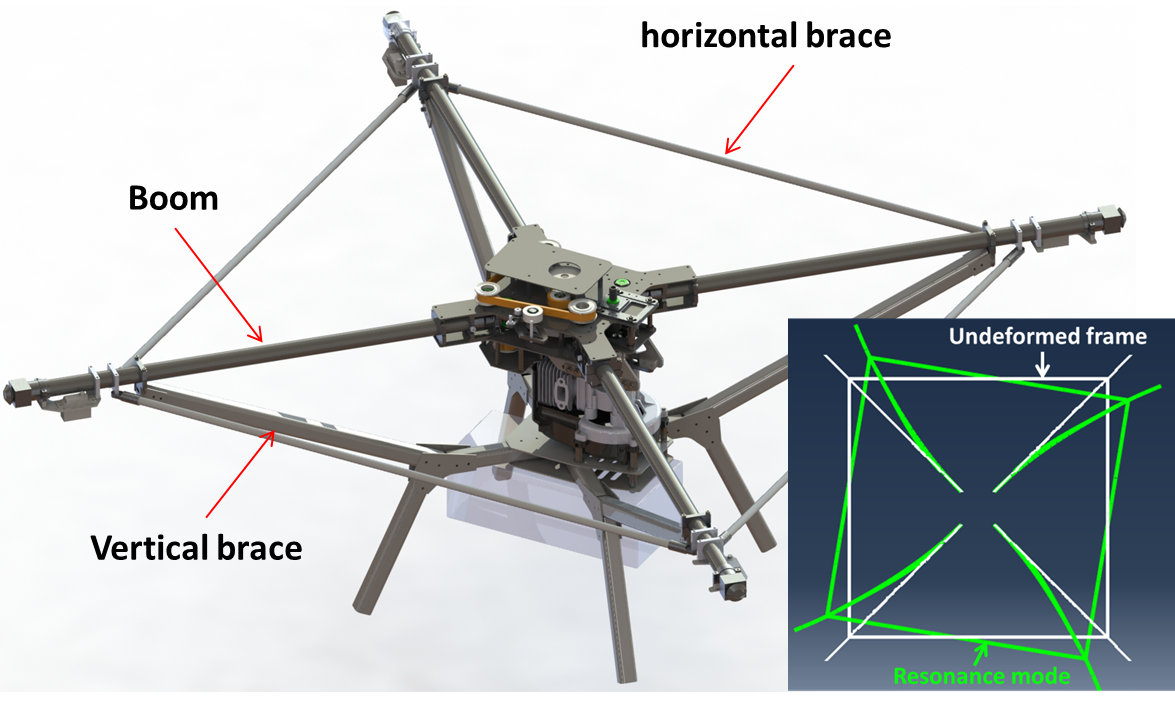}
		\caption{Second version airframe, lower left corner shows the top view of its first resonance mode}
		\label{fig:frame_with_braces}
	\end{figure}

	Our airframe design is shown in Figure \ref{fig:frame_with_braces}. Both vertical and horizontal braces are added to the conventional X-frame to form a truss structure. The vertical braces are intended to improve the booms' rigidity under vertical bending, and the horizontal braces are meant to reduce the booms' horizontal wagging. 
	
	Without the horizontal braces, each boom combined with its vertical brace is independent from the other booms and hence can be analyzed alone. Finite element analysis (FEA) reveals that each boom-brace combination's first resonance mode occurs at 45.3 Hz in the horizontal direction, which is dangerous because it is below the n-per-revolution frequency of the rotors. The addition of horizontal braces may at first seem unhelpful because the first resonance mode of the entire frame still occurs at 42.3 Hz, also in the horizontal direction. However, this resonance mode will never be activated in the context of quadrotors. This is because the resonance mode requires excitations from all rotors to have the same phase to become active (lower left corner of Figure \ref{fig:frame_with_braces}), whereas alternate rotors of quadrotors are contra-rotating, giving out-of-phase excitations that cancel each other. 
	
	The aforementioned conclusions drawn by FEA were also verified experimentally. Before installing the horizontal braces, severe horizontal wagging movement of the booms were observed when the quadrotor was on the ground while the engine was accelerating to its operating speed. In contrast, the wagging movement disappeared after the addition of horizontal braces.

	\section{Flight Control\label{sec.flight control}}
	\subsection{Control Structure \label{subsec.control_structure}}
\begin{figure}[htp]
		\centering
		\includegraphics[width=85mm]{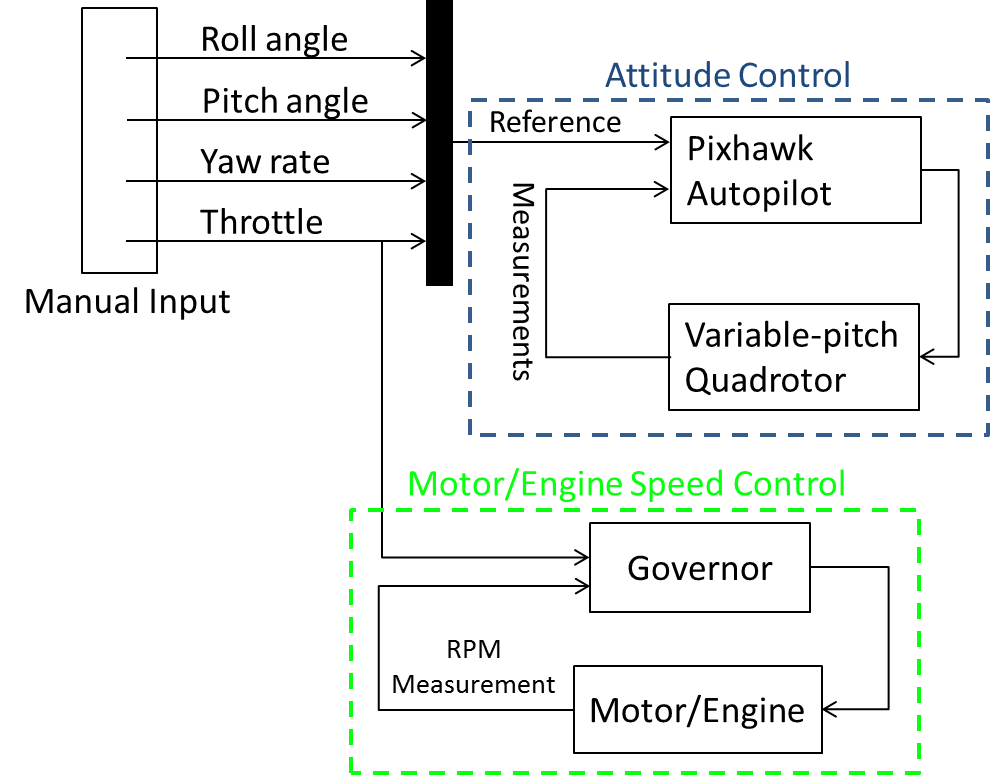}
		\caption{Control structure and implementation}
		\label{fig:control_structure}
	\end{figure}

	Control of the gasoline quadrotor consists of two independent loops. First, attitude stabilization and tracking is realized using a simple PD controller. Second, rotational speed of the rotors is kept constant in a separate loop. The overall control structure is shown in Figure \ref{fig:control_structure}.
	
	A PD controller is chosen for the following reasons. Firstly, only manual control is needed to verify design concepts at the current stage of development. Secondly, the flight envelope is mostly conservative, with the quadrotor flying in hover or near-hover conditions. Lastly, variable-pitch quadrotors' linear relationship between pitch angle and rotor lift/torque resembles conventional quadrotors' quadratic relationship between motor RPM and rotor lift/torque. Therefore, variable-pitch quadrotors can be controlled in a way similar to conventional quadrotors.
	
	For simplicity, it is assumed that the roll, pitch and yaw channels are decoupled second-order systems, which is valid given our operating conditions. The PD gains are calculated using a damping ratio of 0.8 and a natural frequency of 7 rad/s (1.11 Hz). The gains implemented on the quadrotor are summarized in Table \ref{tab:PD_gains}.
	\begin{table}[h]
		\caption{PD controller gains}
		\label{tab:PD_gains}
		\begin{center}
			\begin{tabular}{cccc}
				\hline
				Channel & Moment of Inertia ($kgm^2$)& $K_p$ & $K_d$\\
				\hline
				Roll & 0.43 & 21.1 & 4.8 \\
				Pitch & 0.43 & 21.1 & 4.8 \\
				Yaw & 0.67 & 32.8 & 7.5 \\
				\hline
			\end{tabular}
		\end{center}
	\end{table}

	\subsection{Control Implementation \label{subsec.Control_Implementation}}	
	The PD controller for attitude control is implemented using the Pixhawk autopilot, which also offers more advanced features that will become useful for future development, such as GPS-aided position control. In addition, Pixhawk’s firmware is open source, allowing the implementation of different control techniques when necessary \cite{Pixhawk}. 
	
	Rotor speed is controlled by an engine governor, a common hobbyist device to control the rotational speed of motors or engines. Based on throttle input, the governor generates a user-defined rotor speed reference which is set as a constant value for our quadrotor prototypes. 
	
	\section{Flight Tests\label{sec.flight tests}}
	As mentioned in Section \ref{sec.overview}, the first set of tests was conducted on the electric prototype to verify key aspects of the design methodology. The electric prototype successfully took off in its maiden flight. The take-off weight was 6.3 kg including 2 kg of battery. The quadrotor took off at a pitch angle of $6^{\circ}$ at 2500 RPM, which is in accordance with the lift calculation in Figure \ref{fig:lift}.
	
	Reliability of the entire electric prototype, especially the drivetrain and airframe, has also been tested by running the quadrotor at the hovering pitch angle and at 2500 RPM for an extended period of time. To prevent unexpected accidents from damaging the prototypes, the test was conducted on the ground by strapping additional ballast to the quadrotor frame to prevent it from taking off. The entire system turned out to be reliable enough to survive several hours of testing.

	\begin{figure}[htp]
		\centering
		\includegraphics[width=85mm]{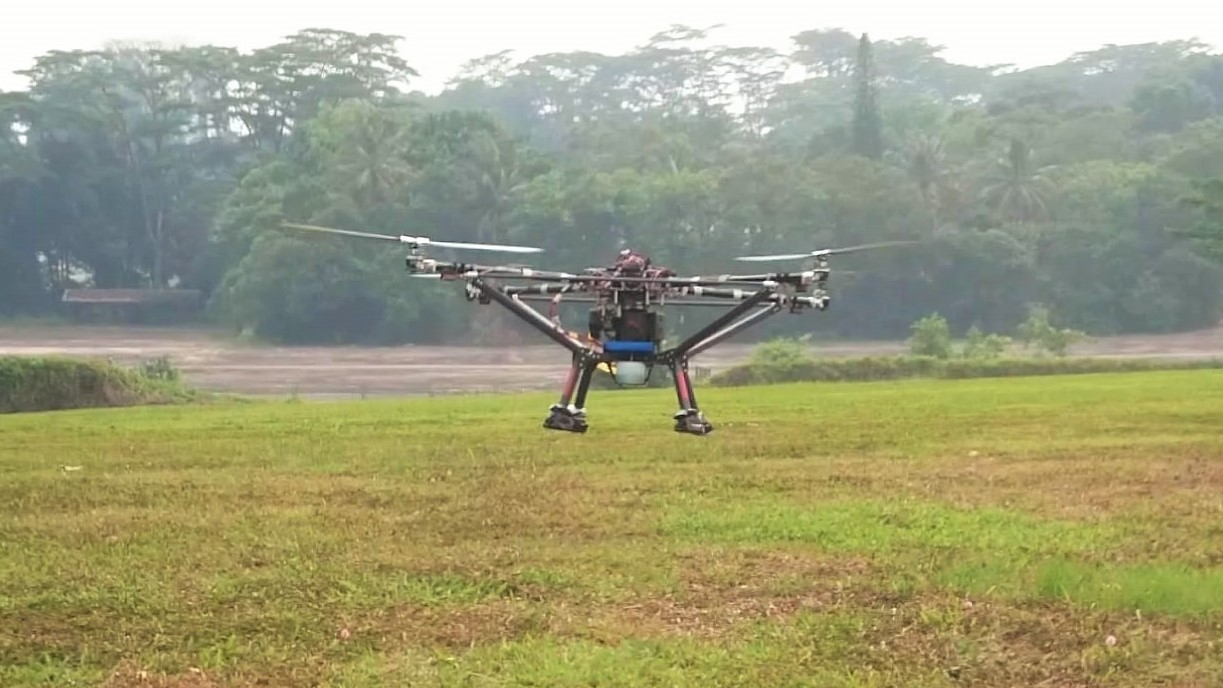}
		\caption{Gasoline quadrotor in flight}
		\label{fig:in_flight}
	\end{figure}
	
\begin{figure}[htp]
		\centering
		\includegraphics[width=88mm]{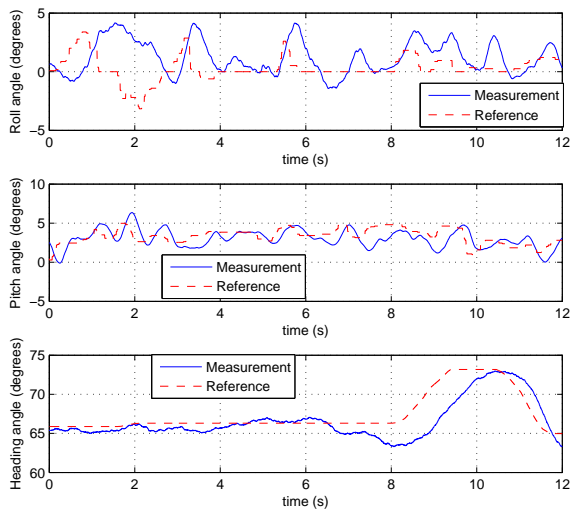}
		\caption{Flight data samples of gasoline engine quadrotor}
		\label{fig:flight_data}
	\end{figure}

	The gasoline prototype has also successfully taken off and hovered for a few minutes with manual control (Figure \ref{fig:in_flight}). The roll, pitch and yaw angles were able to follow the reference input provided by the human pilot, as shown in Figure \ref{fig:flight_data}. However, the designed endurance was not experimentally verified due to various problems manifested during the test. For instance, the operating RPM was lowered during the test because vibration was still too severe. Additional tests and improvements are necessary to achieve long-endurance flight.   

	\section{Conclusion and Future Work \label{sec.concl}}
	Illustrated by our prototypes, a methodology to design and construct variable-pitch long-endurance gasoline-engine quadrotors has been presented. The methodology is comprised of three steps. Firstly, rotor and engine sizes are determined using experimentally-validated results from aerodynamics. The second step is to design and build a reliable drivetrain and airframe. A drivetrain with alternate rotors contra-rotating was creatively realized using a combination of timing belts and gears. On the other hand, vibration turned out to be a prominent concern in airframe design, but can be mitigated with the help of FEA. Lastly, attitude and rotor speed control are implemented respectively using Pixhawk autopilot and engine governor. The methodology was first verified by the successful flight of the proof-of-concept electric quadrotor prototype. The gasoline prototype has also been designed, built and flown with manual control. 
		
	The prototypes have a designed maximum thrust of 16 kg and a maximum take-off weight of 10 kg. Expected flight endurance of the gasoline prototype is 2.8 hours. 
	
	We will continue to test and improve the gasoline prototype to achieve its designed flight endurance. The drivetrain and airframe will also be further optimized to make the quadrotor lighter, more efficient and more reliable. More advanced control techniques could also be implemented to fully utilize the variable-pitch rotors. Moreover, once the 2.8 hour flight endurance is proven, it will greatly expand quadrotors' applications and open up new problems and challenges.

	\section*{Acknowledgment}
	The authors would like to thank Mr. Duojiing Goh, Mr. Joe Hwee and Mr. Kangli Wang for their advice and support during the development and testing of the long-endurance quadrotor prototypes. 

	\bibliographystyle{IEEEtran}
	\bibliography{reference}
	
\end{document}